\documentclass[10pt,twocolumn,letterpaper]{article}

\usepackage{iccv}
\usepackage{times}
\usepackage{epsfig}
\usepackage{graphicx}
\usepackage{amsmath}
\usepackage{amssymb}
\usepackage{bm}
\usepackage{stfloats}
\usepackage{multirow}
\usepackage{varwidth}
\usepackage{capt-of}

\usepackage[breaklinks=true,bookmarks=false]{hyperref}

\iccvfinalcopy 

\def\iccvPaperID{3430} 

\ificcvfinal\fi

\begin{document}

\title{Fast Object Detection in Compressed Video}

\author{Shiyao Wang\textsuperscript{1,2} \thanks{The work was done when Shiyao Wang  was at Tsinghua University.}  \hspace{4mm}  Hongchao Lu\textsuperscript{1} \hspace{4mm}  Zhidong Deng\textsuperscript{1}\\
Department of Computer Science and Technology, Tsinghua University\textsuperscript{1}\\
Alibaba Group\textsuperscript{2}\\
{\tt\small wangshy31@gmail.com  luhc15@mails.tsinghua.edu.cn michael@tsinghua.edu.cn }
}

\maketitle
\ificcvfinal\thispagestyle{empty}\fi

\begin{abstract}
Object detection in videos has drawn increasing attention since it is more practical in real scenarios. 
Most of the deep learning methods use CNNs to process each decoded frame in a video stream individually. However, the free of charge yet valuable motion information already embedded in the video compression format is usually overlooked.
In this paper, we propose a fast object detection method by taking advantage of this with a novel Motion aided Memory Network (MMNet). The MMNet has two major advantages: 1) It significantly accelerates the procedure of feature extraction for compressed videos. It only need to run a complete recognition network for I-frames, i.e. a few reference frames in a video, and it produces the features for the following P frames (predictive frames) with a light weight memory network, which runs fast; 2) Unlike existing methods that establish an additional network to model motion of frames, we take full advantage of both motion vectors and residual errors that are freely available in video streams. To our best knowledge, the MMNet is the first work that investigates a deep convolutional detector on compressed videos. Our method is evaluated on the large-scale ImageNet VID dataset, and the results show that it is  3$\times$ times faster than single image detector R-FCN and 10$\times$ times faster than high-performance detector MANet at a minor accuracy loss.
\end{abstract}

\section{Introduction}
\footnotetext[1]{State Key Laboratory of Intelligent Technology and Systems, THUAI, BNRist, Center for Intelligent Connected Vehicles and Transportation, Tsinghua University, Beijing, China.}

\begin{figure}[t]
\begin{center}
\includegraphics[width=1\linewidth]{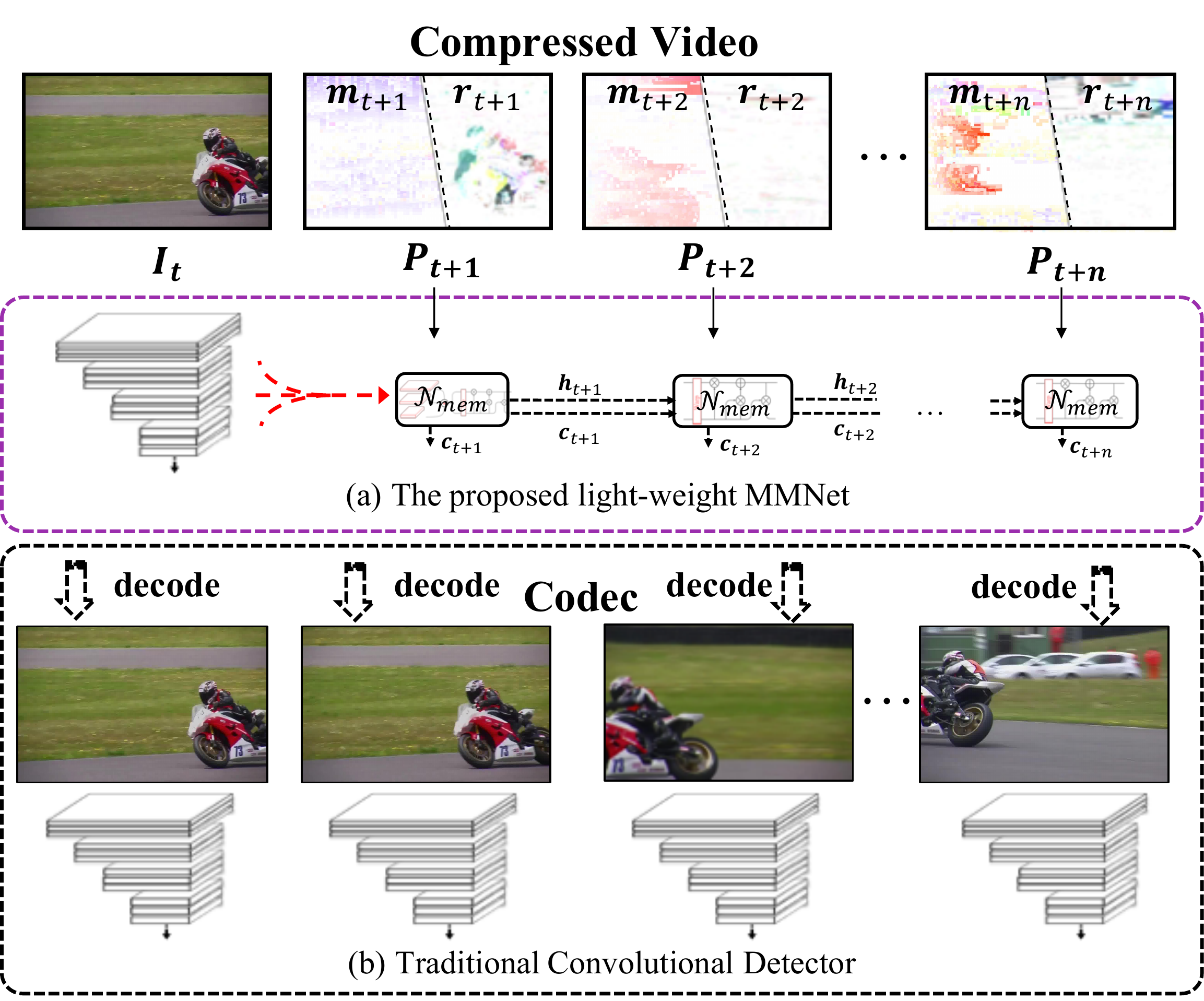}
\end{center}
\caption{(a) The proposed light-weight MMNet accelerates the CNN inference by using compressed video directly, while (b) most previous methods for video analysis use heavy computational networks to extract features frame by frame.}
\label{fig1}
\end{figure}

Video is viewed as one of the next frontiers in computer vision since many real-world data sources are video based, ranging from visual surveillance \cite{bramberger2006distributed}, human-computer interaction \cite{nagi2011max} to autonomous driving \cite{wu2017squeezedet}.
In the past five years, deep learning methods have made historic progress in still image analysis \cite{simonyan2014very,szegedy2016rethinking,he2016deep,Inceptionv4,zoph2018learning}. Novel CNN based frameworks have been proposed for single image object detection, including Faster R-CNN \cite{ren2015faster}, R-FCN \cite{dai2016r}, SSD \cite{liu2016ssd}, YOLO \cite{redmon2016you} and FPN \cite{FPN17}. Although there has great success in static image object detection, it still remains a challenging problem for detection in videos. Since frames may suffer from imaging-related degradations, most previous works \cite{zhu17fgfa,wang2018fully,xiao2018video,kang2017tpn,kang2017t} focus on improving frame-wise detection results. They extract features of the dense frames by applying existing image recognition networks (e.g., ResNet\cite{he2016deep}) individually (see the bottom line in Figure \ref{fig1}), and leverage temporal coherence by feature aggregation or bounding box rescoring. 
Although these methods improve final performance, using CNNs to process the dense frames of videos is computationally expensive while it becomes unaffordable as the video goes longer. 

In order to reduce the redundant computation, \cite{zhu17dff,zhu2018towardsMobiles} propose methods that run an expensive feature extractor only on sparse keyframes and then propagate the resulting deep features to other frames. The key idea for feature propagation among frames is to calculate pixel-wise displacements through FlowNet\cite{dosovitskiy2015flownet}. However, it pays extra time for calculating displacements since the FlowNet still composes dozens of convolutional layers. They treat a video as a sequence of independent images and ignore the fact that a video is generally stored and transmitted in a compressed data format. The codecs split a video into I-frames (intra-coded frames) and P/B-frames (predictive frames). An I‑frame is a complete image while a P/B-frame only holds the changes compared to the reference frame. For example, the encoder stores the object's movements $\bm{m}_{t+k}$ and residual errors $\bm{r}_{t+k}$ when it moves across a stationary background (see the top line in Figure \ref{fig1}). So, consecutive frames are highly correlated, and the changes are already encoded in a video stream. Treating them as a sequence of still images and exploiting different techniques to retrieve motion cues seem time-consuming and cumbersome.

In this paper, we propose a fast and accurate object detection method for compressed videos called Motion-aided Memory Network with pyramidal feature attention (MMNet). For a group of successive pictures (GOP) in a video, it runs the complete recognition network for I-frames, while a light-weight memory is developed to produce features for the following P-frames. The proposed MMNet receives the features of the preceding I-frame as input and fast predicts the following features by using motion vectors and residual errors in video streams. Moreover, different from the previous work that only propagates high-level features, the proposed memory network composes pyramidal features which enable the model to detect objects across multiple scales. In summary, the contributions of this paper include:

- We explore inherent motion signals and residual errors in codecs to align and refine features. Note that the signals retain necessary motion cues and are freely available.
 
- We propose a pyramidal feature attention that enables the memory network to propagate features from multiple scales. It helps to detect objects across different scales.
   
- We evaluate the proposed model on the large-scale ImageNet VID dataset \cite{russakovsky2015imagenet} and present memory visualization for further analysis. Our model achieves significant speedup at a minor accuracy loss.

\section{Related work}
\subsection{Object detection}
\subsubsection{Object detection from still images}
State-of-the-art methods for general object detection consist of feature networks \cite{krizhevsky2012imagenet,simonyan2014very,szegedy2015going,he2016deep,Inceptionv4,dense,dpn} and detection networks \cite{rcnn2014,girshick2015fast,spp15,ren2015faster,dai2016r,FPN17,ohem,focal}. \cite{rcnn2014} is a typical proposal based detector which uses extracted proposals \cite{selective2013}. Faster R-CNN \cite{ren2015faster} further integrates proposal generation step into CNNs. R-FCN \cite{dai2016r} has comparable performance and higher speed compared to Faster R-CNN. 
We use R-FCN as our baseline and its computation speed is further improved for video object detection.

\subsubsection{Object detection in videos}
One of the main-stream methods is based on per-frame complete detection and improves the detection quality by leveraging temporal coherence. And the other tries to speed up the computation by using temporal redundancy.

\textbf{For high performance}, \cite{zhu17fgfa,kang2017tpn,wang2018fully,xiao2018video,zhu2018towards,bertasius2018object} propose end-to-end learning models to enhance per-frame features. \cite{zhu17fgfa,wang2018fully,zhu2018towards} adopt FlowNet \cite{dosovitskiy2015flownet} to align and aggregate features. 
\cite{kang2017tpn} provides a novel tubelet proposal network to efficiently generate spatiotemporal proposals. \cite{xiao2018video} computes the correlation between neighboring frames and introduces a memory module to aggregate their features. \cite{bertasius2018object} uses deformable convolutions across time to align the features from the adjacent frames. \cite{bertasius2018object,kang2017t,feichtenhofer2017detect} are based on detected bounding boxes rather than feature-level aggregation. \cite{han2016seq,kang2016object,kang2017t} propose mapping strategies of linking still image detections to cross-frame box sequences in order to boost scores of weaker detections.  D\&T \cite{feichtenhofer2017detect} is the first work to joint learn ROI tracker along with detector and the tracker is also exploited to link the cross-frame boxes. All of these mentioned works achieve high detection performance but they use a computationally expensive network to generate per-frame features.

\begin{figure*}[t]
\begin{center}
\includegraphics[width=1\linewidth]{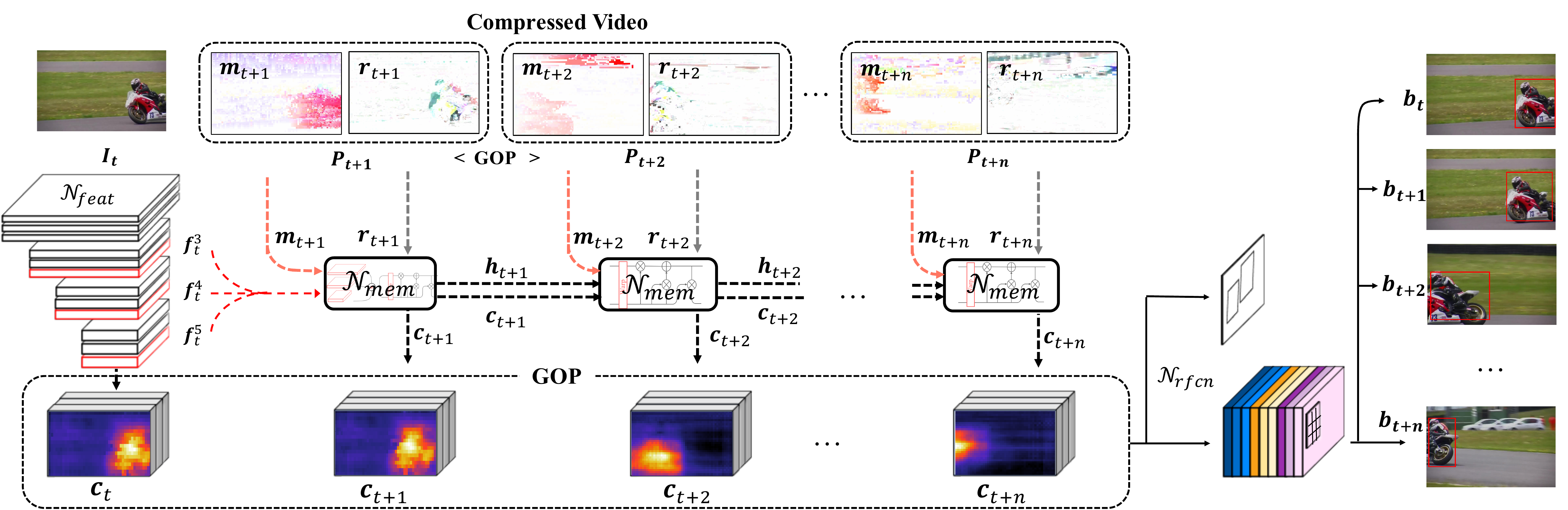}
\end{center}
   \caption{The overall framework of the proposed MMNet with pyramidal feature attention. The feature extractor $\mathcal{N}_{feat}$ only runs on reference frame $\bm{I}_t$, and the other features of $\bm{c}_{t+k}$ are generated by the memory network $\mathcal{N}_{mem}$. Motion vectors $\bm{m}_{t+k}$ and residual errors $\bm{r}_{t+k}$ are fed into the memory network so as to provide motion cues. Finally, all the features in one GOP (group of pictures) are aggregated to the detection network $\mathcal{N}_{rfcn}$, producing bounding boxes simultaneously.}
\label{fig2}
\end{figure*}

\textbf{For fast inference},
\cite{zhu17dff} utilizes optical flow network for calculating pixel-level correspondence and propagating deep feature maps from keyframes to other frames.  The flow estimation and feature propagation are faster than feature networks.
Thus, the significant speedup is achieved.\cite{chen2018optimizing} introduces temporal propagation on box-level. They first produce bounding boxes on keyframes, and then generate boxes of other frames through a coarse-to-fine network. \cite{liu2017mobile} propagates feature maps across frames via a convolutional LSTM. They only use appearance features without explicitly capturing motion cues. Although their model is faster than existing methods, the performance is much degraded. \cite{wang2016accelerating,howard2017mobilenets,iandola2016squeezenet} also focus on model accelerating. They aim to build light-weight deep neural networks which are unrelated to specific tasks.

\subsection{Deep learning model on compressed videos}
H.264/MPEG-4 Part 10, Advanced Video Coding \cite{Sofokleous05} is one of the most commonly used formats for recording, compression and distribution of videos. It is a block-oriented motion-compensation-based video compression standard \cite{richardson2003h264}. To our knowledge, only a few prior works applied deep models directly on compressed videos. \cite{KantorovL14,ToreyinCAA05} utilize signals from compressed video to produce non-deep features. \cite{wu2018compressed}  resembles our model the most and they aim to improve both speed and performance on video action recognition which focuses on producing video-level features. But the video object detection needs to produce per-frame bounding boxes that has per-frame feature quality requirements. 

\section{Method}
\subsection{Overview}
\label{overview}
The proposed motion-aided memory network with pyramidal feature attention is presented in Figure \ref{fig2}. 

\emph{For the input video}, we use H.264 baseline profile 
as illustration since these compression techniques that leverage consecutive frames are usually similar. H.264 baseline profile contains two types of frames: I- and P-frames. An I-frame (denoted as $\bm{I}_t \in \mathbb{R}^{h\times w\times 3}$) is a complete image. $h$ and $w$ are the height and width. P-frames are also known as delta‑frames, denoted as $\bm{P}_{t+k}$. They can be reconstructed by using the stored offsets, called motion vectors $\bm{m}_{t+k}$ and residual errors $\bm{r}_{t+k}$. Detailed illustration of extracting $\bm{m}_{t+k}$ and $\bm{r}_{t+k}$ is presented in Section \ref{motion}. In Figure \ref{fig2}, we show a typical GOP on the top line, denoted as \{$\bm{I}_t, \bm{P}_{t+1}, \cdots, \bm{P}_{t+k},\cdots, \bm{P}_{t+n}$\}. 

\emph{For the core modules}, there are three networks: feature extractor, memory network and detection network, indicated as $\mathcal{N}_{feat}$, $\mathcal{N}_{mem}$ and $\mathcal{N}_{rfcn}$, respectively. The I-frame $\bm{I}_t$ is fed to the $\mathcal{N}_{feat}$ in order to generate pyramidal features $\bm{f}_t^l \in \mathbb{R}^{h^l\times w^l \times c^l}$. $l$ is the index of multiple stages in a network and $w^l, h^l$ and $c^l$ are the corresponding width, height and channel numbers. They are sent to the memory network $\mathcal{N}_{mem}$ so as to fast produce features of the following frames $[\bm{c}_{t+1}, \cdots, \bm{c}_{t+n}]$. The memory network contains two modules: pyramidal feature attention $\mathcal{N}_{atten}$ (in Section \ref{pyramid}) and motion-aided LSTM $\mathcal{N}_{m-lstm}$ (in Section \ref{lstm}). Pyramidal feature attention receives $\bm{f}_t^l$ as input and generate $\bm{f}_{t}^{atten}$ that will be propagated to the neighboring frames. And the motion-aided LSTM transfers the preceding features by using motion vectors $\bm{m}_{t+k}$ and residual errors $\bm{r}_{t+k}$.  The above procedure is formulated as:
\begin{equation}\label{equ1}
\setlength\abovedisplayskip{0pt}
\setlength\belowdisplayskip{0pt}
\begin{split}
\bm{f}_t^l = \mathcal{N}_{feat}(\bm{I}_{t})\\
\end{split}
\end{equation} 
\begin{equation}
\bm{c}_{t+k}=\left\{
\begin{array}{rcl}
\mathcal{N}_{atten}(\bm{f}_{t}^{3}, \bm{f}_{t}^{4}, \bm{f}_{t}^{5})  & {k=0}\\
\mathcal{N}_{m-lstm}(\bm{c}_{t+k-1}, \bm{m}_{t+k}, \bm{r}_{t+k}) & {1\le k \le n}\\
\end{array} \right.
\end{equation}
\begin{equation}\label{equ3}
\begin{split}
[b_t, b_{t+1},\cdots, b_{t+n}] = \mathcal{N}_{rfcn}([\bm{c}_t, \bm{c}_{t+1}, \cdots, \bm{c}_{t+n}]) \\
\end{split}
\end{equation} 
where $[\bm{c}_t, \bm{c}_{t+1}, \cdots, \bm{c}_{t+n}]$ denotes the concatenation of the features of one GOP. It means that $\mathcal{N}_{rfcn}$ will receive the features within the same GOP, and predict their bounding boxes $[b_t, b_{t+1},\cdots, b_{t+n}] $ simultaneously. 


\begin{figure*}[t]
\begin{center}
\includegraphics[width=1.0\linewidth]{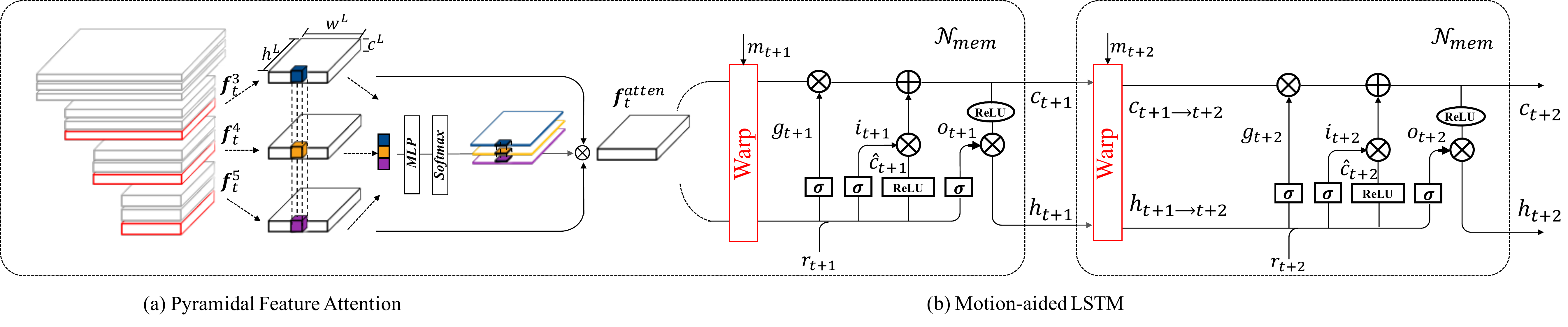}
\end{center}
\vspace{-1em}
   \caption{Light-weight MMNet with pyramidal feature attention. Attention mechanism aims to selectively combine the pyramidal features. Motion vectors are used to calibrate cell/hidden features before they run through the memory module. Residual errors are employed to correct appearance changes.}
\label{fig3}
\end{figure*}

\subsection{Pyramidal Feature Attention}
\label{pyramid}
Previous methods only propagate high-level feature maps to the neighboring frames (``res5c$\_$relu'' in \cite{zhu17dff}). In other words, the following P-frames only receive the high-level semantic features from the reference frame. It is not friendly for detecting objects at vastly different scales. \cite{FPN17} utilizes the inherent pyramidal hierarchy of deep CNNs to detect multi-scale objects. However, their predictions are independently made on each level. If we employ the method from still images to videos, we should propagate features and make predictions on several scales at each timestamp.  On the contrary, we develop a method that adaptively combines the pyramidal features through attention mechanism within the first memory module (see the first block in Figure \ref{fig3}). The combined pyramidal features are then sent to the motion-aided LSTM (see Section \ref{lstm}).

First, we gather the pyramidal features $\bm{f}^l_t$ from different stages.  A network can be divided into $L$ stages whose layers produce output maps of the same resolution. We define one pyramid level for each stage and use the later stages $l=3,4,5$ of a network(see Figure \ref{fig3}(a)). We utilize \emph{\{res3b3\_relu, res4b22\_relu, res5c\_relu\}} of ResNet-101 as the input and transform them into the same dimension:
\begin{equation}\label{sec3.2-eq1}
\setlength\abovedisplayskip{4pt}
\setlength\belowdisplayskip{4pt}
\begin{split}
\widehat{\bm{f}}_t^l = \mathcal{F}_{embed}(\bm{f}_t^l, \bm{f}_t^L), \forall l, 3\le l \le L
\end{split}
\end{equation}
where $\mathcal{F}_{embed}(\cdot)$ can be implemented as a convolution layer with the proper stride. The target dimension is the size of $\bm{f}_t^L$. Hence, $\widehat{\bm{f}}_t^l$ with different $l$ have the same dimension which is necessary for the following attention operation.

Second, we use squeeze operation across the channel axis to represent features of each scale $l$ at position $(i, j)$:
\begin{equation}\label{sec3.2-eq2}
\setlength\abovedisplayskip{4pt}
\setlength\belowdisplayskip{4pt}
\begin{split}
\emph{e}_t^l(i, j) = \sum_{k=1}^{c^L} \widehat{\bm{f}}_t^l(i, j, k),\\
\forall i,j, 1\le i \le w^L, 1\le j \le h^L\\
\end{split}
\end{equation} 
where $i$ and $j$ enumerate all spatial locations in the feature maps. The squeeze operation sums all the elements
across the channel dimension which can be viewed as feature salience. We call the above outputs ``scale descriptors''. It is inspired by SENet \cite{hu2018squeeze}, but they use global average pooling to collect statistics for the spatial dimension. 

Finally, we adopt the scale descriptors as input to generate attention weights in order to adaptively combine the features from different scales. We define the fused representation $\bm{f}_t^{atten}$ and attention weights $\alpha_t^l(i, j)$ as follows:

\begin{equation}\label{sec3.3.1-eq3}
\setlength\abovedisplayskip{4pt}
\setlength\belowdisplayskip{4pt}
\begin{split}
\bm{f}_t^{atten}(i, j) &= \sum_{l=3}^{L}{\alpha_t^l(i, j)\widehat{\bm{f}}_t^l(i, j)}\\
\sum_{l=3}^L \alpha_t^l(i, j) &= 1 \\
\end{split}
\end{equation}
The attention weights are produced as follows:
\begin{equation}\label{sec3.2-eq4}
\setlength\abovedisplayskip{4pt}
\setlength\belowdisplayskip{4pt}
\begin{split}
\bm{\alpha}_t(i, j) &=softmax(\mathbf{MLP}(\bm{e}_t(i, j))) \\
\bm{e}_t(i, j) &= [\emph{e}_t^3(i, j), \emph{e}_t^4(i, j), \emph{e}_t^5(i, j)]\\
\bm{\alpha}_t(i, j) &= [\alpha_t^3(i, j), \alpha_t^4(i, j), \alpha_t^5(i, j)]\\
\end{split}\end{equation}

After being processed by the pyramidal feature attention, $\bm{f}_t^{atten}$ will be fed into the motion-aligned LSTM.

\subsection{Motion-aided Memory Network}
\subsubsection{Motion Vectors and Residual Errors}
\label{motion}

\begin{figure}[h]
\vspace{-0.5em}
\begin{center}
\includegraphics[width=1.0\linewidth]{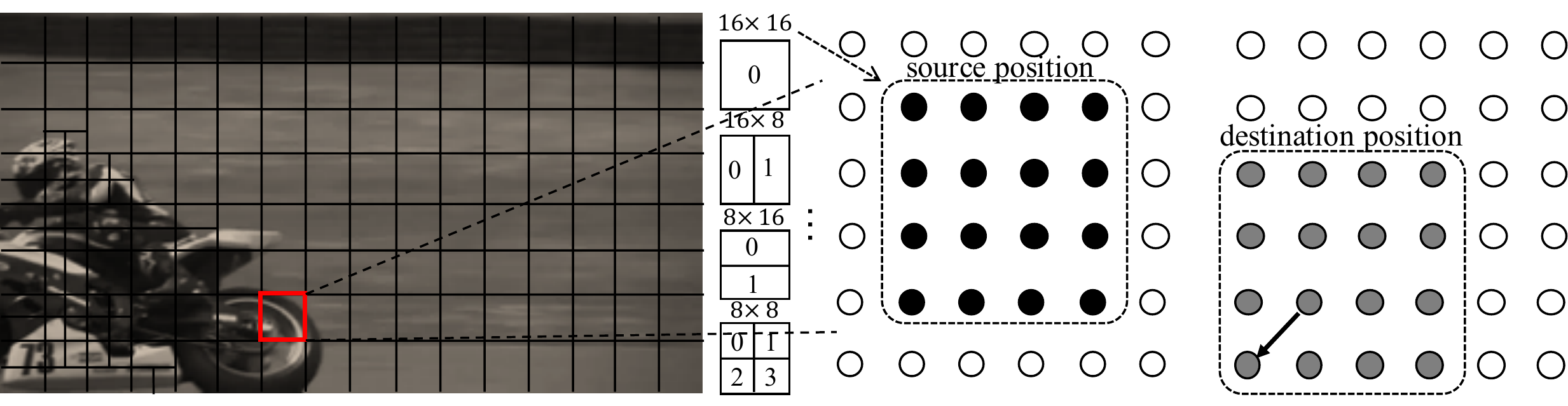}
\end{center}
   \caption{Motion vector represents a macroblock in a picture based on the position of that in another picture.}
\label{mv}
\end{figure}

For compressed data, a P-frame is divided into blocks known as macroblocks (see Figure \ref{mv}). The supported prediction block sizes range from $4\times4$ to $16\times16$ samples. Video encoder adopts a block matching algorithm \cite{zhu2000new,nie2002adaptive,yang2006block,je2013optimized} to find a block similar to the one it is encoding on a previously encoded frame. The absolute source position and destination position of a macroblock are stored in the motion vectors. Moreover, if there is not an exact match to the block it is encoding, the residual errors are also sent to the decoder. 
\vspace{-0.5em}

In our implementation, we use FFmpeg \cite{ffmpeg} to extract the motion vectors and residual errors for each P-frame. 
When we obtain the original motion vectors and residuals from the codecs, we resize them to match the size of the feature maps $h^L$ and $w^L$. And the motion vectors should be further rescaled by spatial stride since the original values indicate the movements in the decoded frames.

\subsubsection{Motion-aided LSTM}
\label{lstm}

We use LSTM \cite{HochreiterS97} to transfer the features. There are two modifications to the conventional LSTMs. One is the motion vector aided feature warping and another is residual error based new input.

Although the gates in LSTM focus on selecting and updating representations, it is still hard for them to forget the object after it has moved to a different position \cite{xiao2018video}. Experiments in Section \ref{sec:vis:1} demonstrate this concern. It is called misaligned features across frames. Hence, we propose a motion vector based feature warping which helps to calibrate cell/hidden features before running through the memory module (see Figure \ref{fig3}(b)). We warp the feature maps from the neighboring frames to the current frame as follows:
\begin{equation}\label{equ4}
\setlength\abovedisplayskip{4pt}
\setlength\belowdisplayskip{4pt}
\begin{split}
\bm{c}_{t+k-1 \to t+k} &= \mathcal{W}(\bm{c}_{t+k-1}, \bm{m}_{t+k})\\
\bm{h}_{t+k-1 \to t+k} &= \mathcal{W}(\bm{h}_{t+k-1}, \bm{m}_{t+k})
\end{split}
\end{equation} 
where $\bm{c}_{t+k-1}$ and $\bm{h}_{t+k-1}$ are outputs of the memory module at time $t+k-1$. We set $\bm{c}_{t}$ and $\bm{h}_{t}$ to $\bm{f}_{t}^{atten}$ and $k \in [1, n]$. $n$ is the number of P-frames in a GOP. The warping operation $\mathcal{W}$ is similar to \cite{zhu17fgfa}. It is implemented by bi-linear function which is applied on each location for all feature maps. It projects a location $\bm{p}+\Delta \bm{p}$ in the frame $t+k-1$ to the location $\bm{p}$ in the frame $t+k$ which can be formulated as:
\begin{equation}\label{equ5}
\setlength\abovedisplayskip{4pt}
\setlength\belowdisplayskip{4pt}
\begin{split}
\Delta \bm{p} &=  \bm{m}_{t+k}(\bm{p})\\
\bm{c}_{t+k-1 \to t+k}(\bm{p}) &= \sum_{\bm{q}}{G(\bm{q}, \bm{p} + \Delta \bm{p})\bm{c}_{t+k-1}(\bm{q})}\\
\end{split}
\end{equation}
where $\Delta \bm{p}$ is obtained through $\bm{m}_{t+k}$. $\bm{q}$ enumerates all spatial locations in the feature maps $\bm{c}_{t+k-1}$, and $G(\cdot)$ denotes bi-linear interpolation kernel as follow:
\begin{equation}\label{equ6}
\setlength\abovedisplayskip{4pt}
\setlength\belowdisplayskip{4pt}
\begin{split}
G(\bm{q}, \bm{p} + \Delta \bm{p}) = max(0, 1 - ||\bm{q}  - (\bm{p} + \Delta \bm{p})||)\\
\end{split}
\end{equation}
Hidden features $\bm{h}_{t+k-1 \to t+k}$ can also be obtained through above operations. Then $\bm{c}_{t+k-1 \to t+k}$ and $\bm{h}_{t+k-1 \to t+k}$ are used as the input from previous time to the current memory module. 

For conventional LSTMs, the current complete frame will be used as the new information. In our model, we use the residual errors as new input. Through the motion vector, the previous features can be matched to the current state, but the current representation still lacks some information. So the video encoder computes the residual errors, whose values are known as the prediction error and needed to be transformed and sent to the decoder. After spatial alignment, the residual errors can be used as the complementary information which are more crucial than the whole appearance features of the complete image. In order to better match the residual errors from image-level to the feature-level, we use one convolutional layer to rescale the values.

After obtaining the warped features and new input, the memory can generate the new cell features as follow:
\begin{equation}\label{equ7}
\setlength\abovedisplayskip{4pt}
\setlength\belowdisplayskip{4pt}
\begin{split}
\bm{g}_{t+k} &= \sigma(\bm{W}_g(\bm{h}_{t+k-1 \to t+k}, \bm{r}_{t+k})), \\
\bm{i}_{t+k} &= \sigma(\bm{W}_i(\bm{h}_{t+k-1 \to t+k}, \bm{r}_{t+k})), \\
\hat{\bm{c}}_{t+k} &= ReLU(\bm{W}_c(\bm{h}_{t+k-1 \to t+k}, \bm{r}_{t+k})), \\
\bm{c}_{t+k} &= \bm{g}_{t+k} \otimes \bm{c}_{t+k-1 \to t+k} + \bm{i}_{t+k} \otimes \hat{\bm{c}}_{t+k} \\
\end{split}
\end{equation} 
where $\oplus$ and $\otimes$ are element-wise addition and multiplication, and $\bm{W}_g, \bm{W}_i$ and $\bm{W}_c$ are learnable weights. $\bm{g}_{t+k}$ can be regarded as a selection mask and $\hat{\bm{c}}_{t+k}$ is new information that holds complementary representation. $\bm{c}_{t+k}$ represents the current frame that will be fed into $\mathcal{N}_{rfcn}$. Then the hidden features can be generated:
\begin{equation}\label{equ8}
\setlength\abovedisplayskip{4pt}
\setlength\belowdisplayskip{4pt}
\begin{split}
\bm{o}_{t+k} &= \sigma(\bm{W}_o(\bm{h}_{t+k-1 \to t+k}, \bm{r}_{t+k})), \\
\bm{h}_{t+k} &= \bm{o}_{t+k} \otimes ReLU(\bm{c}_{t+k})
\end{split}
\end{equation}

Based on this architecture, we can transform former features to the current state and they will be passed to the next step until encountering another new I-frame. Features of one GOP $[\bm{c}_{t}, \bm{c}_{t+1}, \bm{c}_{t+1}, \cdots, \bm{c}_{t+n}]$ will be sent to the detection network $\mathcal{N}_{rfcn}$, producing bounding boxes of objects simultaneously.

\section{Experiments}
\subsection{Dataset preparation and evaluation metrics}
We evaluate the proposed MMNet on the ImageNet \cite{russakovsky2015imagenet} object detection from video (VID) dataset. It is split into 3862 training and 555 validation videos. It contains 30 classes labeled with ground truth bounding boxes on all frames.  We report the evaluation of previous state-of-the-art models on the validation set and use mean average precision (mAP) as the evaluation metric by following the protocols in \cite{kang2017tpn,zhu17fgfa,zhu17dff}. VID releases both original videos and decoded frames. Note that all of the previous state-of-the-art methods use decoded frames as input. It is the first time to detect objects on the original videos on VID. 

The 30 object categories in ImageNet VID are a subset of 200 categories in the ImageNet DET dataset. We follow previous approaches and train our model on an intersection of ImageNet VID and DET set.

\subsection{Training and Evaluation}
We perform two phrase training: 1) the model is trained on the mixture of DET and VID for 12K iterations, with learning rates of $2.5\times 10^{-4}$ and $2.5 \times 10^{-5}$ in the first 80K and 40K iterations, respectively. We use a batch size of 4 on 4GPUs. 2) the motion-aided memory network is integrated into R-FCN, and trained for another one epoch on VID dataset. In this phase, each GPU holds multiple samples in one GOP. It is already introduced by Section \ref{overview}. The feature extractor ResNet101 model is pre-trained for ImageNet classification as default. In both training and testing, we use single scale images with shorter dimension of 600 pixels. For testing we run the whole recognition network only on I-frame and fast predict the bounding boxes for the rest frames. 

\begin{table*}[t] 
\centering 
\renewcommand\arraystretch{1.4}
\begin{tabular}{p{3cm}|p{1.5cm}|p{1.5cm}|p{1.5cm}|p{1.5cm}|p{1.5cm}|p{1.5cm}} 
\hline
Backbone & \multicolumn{6}{c}{ResNet-101} \\
\hline
Methods & (a)&(b)&(c)& (d)&(e)&(f) \\
\hline
MV?       & & & {${\surd}$} &{${\surd}$} &{${\surd}$}&{${\surd}$} \\
Residual?   &  &{${\surd}$} &{${\surd}$} & & {${\surd}$} &{${\surd}$}\\
LSTM? & {${\surd}$}&{${\surd}$} & &{${\surd}$} &{${\surd}$} &{${\surd}$} \\
Pyramidal Attention? &&&&&&{${\surd}$}\\
\hline
mAP(\%)(fast)  &27.7& 27.3 $\downarrow_{0.4}$&38.5 $\uparrow_{10.8}$&43.1 $\uparrow_{15.4}$&44.2 $\uparrow_{16.5}$&43.7$\uparrow_{16.0}$\\
\hline
mAP(\%)(medium) &68.2& 68.0 $\downarrow_{0.2}$&71.2 $\uparrow_{3.0}$&71.5 $\uparrow_{3.3}$&72.0 $\uparrow_{3.8}$&73.4 $\uparrow_{5.2}$\\
\hline
mAP(\%)(slow) &82.6&82.2 $\downarrow_{0.4}$&83.5 $\uparrow_{0.9}$&83.0 $\uparrow_{0.4}$ &83.6 $\uparrow_{1.0}$&84.7 $\uparrow_{2.1}$\\
\hline
\textbf{mAP(\%)} &\textbf{66.3}& \textbf{66.1 $\downarrow_{0.2}$}&\textbf{70.3 $\uparrow_{4.0}$}&\textbf{71.3$\uparrow_{5.0}$} &\textbf{72.1 $\uparrow_{5.8}$}&\textbf{73.0 $\uparrow_{6.7}$}\\
\hline
Speed(fps) & 42.1& 41.9&51.3 & 41.9&41.7& 40.5 \\
\hline
\end{tabular}
\vspace{0.5em}
\caption{Accuracy of different methods on ImageNet VID validation using ResNet-101 feature extraction networks.}
\label{table1}
\end{table*}

\subsection{Ablation Study}
\label{ablation}
In this section, we conduct an ablation study to prove the effectiveness of the proposed network, including motion vectors, residual errors, LSTM and pyramidal feature attention. We use ResNet-101 to extract I-frame features and adopt different ways to propagate features to the following P-frames. The evaluation protocols follow the previous work \cite{zhu17fgfa}. They divide the ground truth objects into three groups according to their movement speed. They use object' averaged intersection-over-union(IoU) scores with its corresponding instances in the nearby frames as the measurement. The lower motion IoU($<0.7$) indicates the faster movement. Otherwise, the larger Motion IoU ($score > 0.9$) denotes the object moves slowly. Table \ref{table1} also shows the accuracy and runtime speed for the models.

\textbf{\emph{Method (a) and (b)}}: Method(a) adopts LSTMs to transform features. It is a conventional solution and we regard it as our baseline. However, without explicit motion cues, LSTMs are unable to automatically align features from previous frames, leading to poor performance (66.3\%mAP in Table \ref{table1}), especially for fast moving objects (27.7\% mAP). In method (b), without motion alignment, the residual errors even hurt the result (66.1\% mAP).

\textbf{\emph{Method (c) and (d)}}: These methods utilize motion vector to warp(/align) the features. Residual errors or LSTMs aim to learn complementary features. we find that \{motion vector + residual error\} is a practical solution because it has the least computational cost (51.3 fps) with comparable accuracy (70.3\%). For the fast moving objects, the result is improved from 27.7\% to 38.5 \%. It also proves that motion information encoded in the compressed videos is valuable for modeling differences among frames. 

\textbf{\emph{Method (e) and (f)}}: These methods are based on motion-aided memory network with/without pyramidal feature attention. Method (e) only propagates high-level feature maps of the top layer and Method (d) delivers pyramidal features to the memory network. We find pyramidal features can further improve the performance with little higher runtime complexity.

To sum up, the motion vector and residual errors are necessary for modeling the motion among consecutive frames. They can speed up the detection procedure. Besides, LSTM is employed to filter out unnecessary information and complement new information. Moreover, propagation of pyramidal features can further improve the detection accuracy. Consequently, these modules are capable of promoting the final feature representations collaboratively.

\subsection{Visualization}
\label{sec:vis:1}
\begin{figure*}[t]
\begin{center}
\includegraphics[scale=0.35,width=1.0\linewidth]{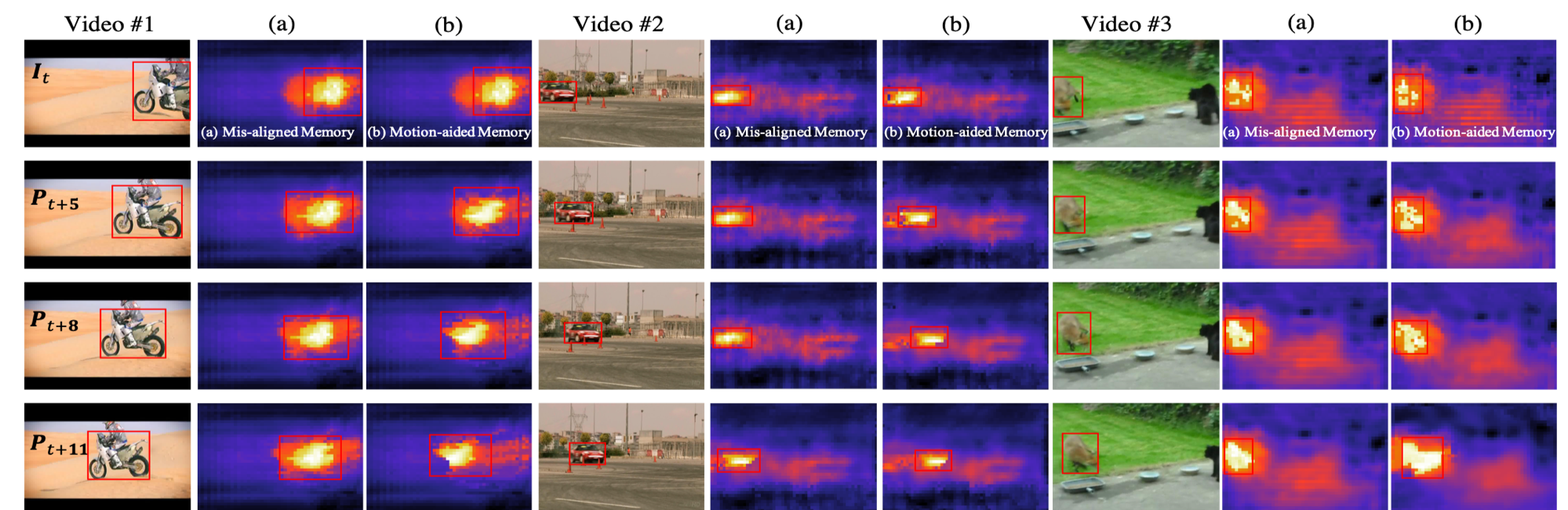} 
\caption{Memory visualization. Each example contains original frames, (a) mis-aligned memory and (b) motion-aided memory. Motion information is quite necessary for feature propagation. It helps MMNet align the feature when the objects move to a different position.} 
\label{fig5} 
\end{center}
\end{figure*}

\textbf{Visualization of Memory}. We attempt to take a deeper look at intermediate features learned by motion-aided memory network. In Figure \ref{fig5}, there are three typical video snippets. For example, in video \#2, the left part consists of decoded frames \{$\bm{I}_t, \cdots, \bm{P}_{t+2}, \cdots, \bm{P}_{t+5}, \cdots, \bm{P}_{t+7}$\}. The car in video moves from left to the middle. We compare the visualization results of mis-aligned and motion-aided memory. 

\begin{figure*}[bp]
 \includegraphics[width=0.995\linewidth]{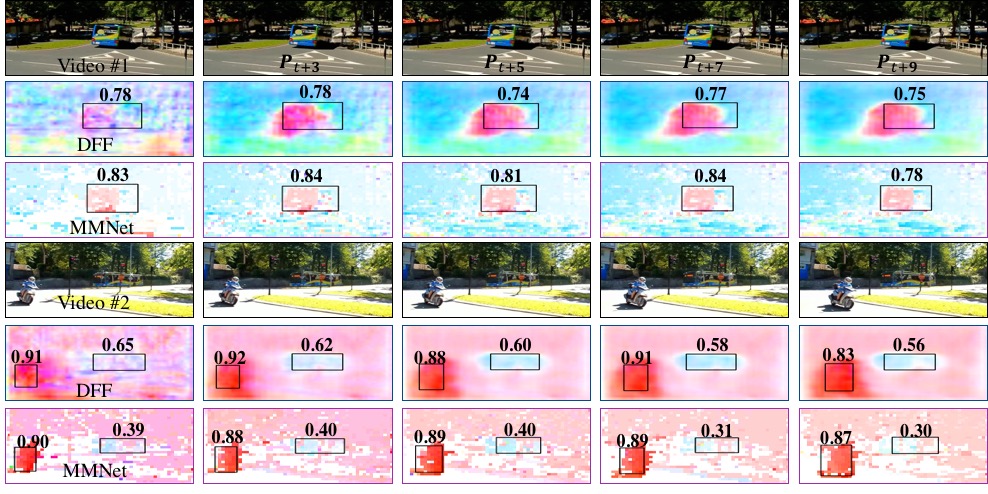}
 \caption{Visualization of FlowNet and Motion Vector. The FlowNet in \cite{zhu17dff} is capable of building detailed model information. And the motion vector can quickly provide the motion cues which helps to speed up detection procedure in most situations.}
\label{fig6}
\end{figure*}

The features in the middle part are learned by LSTM. Although the gates within LSTM are designed for learning the changes among historical information and new input, it is incapable of aligning the spatial features across frames. It cannot capture motion cues only depended on appearance features. The right part presents the motion-aided memory network. Features of \{$\bm{P}_{t+2}, \bm{P}_{t+5}, \bm{P}_{t+7}$\} are all based on $\bm{I}_t$. The MMNet receives codecs information as input, aligns and corrects the propagated features. The neurals of high response in the heapmap move from the left to the middle as the original car. 

From the above comparison, motion information is extremely important for feature propagation. It helps to align features when the objects move to a different position. Thus the motion-aided memory network can calibrate features and alleviate inaccurate localizations.

\textbf{Visualization of FlowNet and Motion Vector}. In order to show the differences of motion cues between the flow estimation and motion vectors, we visualize two examples and their results in Figure \ref{fig6}. Each of the two examples contains the original short snippet, results of FlowNet \cite{dosovitskiy2015flownet,zhu17dff} and motion vectors (We use the tool provided with Sintel\cite{butler2012naturalistic} to visualize the above motion information).

The main advantage of motion vector is freely available. It requires no extra time or models to retrieve motion information because it has already been encoded in the compressed video. From the results in Figure \ref{fig6}, even the motion vector is not as meticulous as FlowNet, it is able to model the motion tendency of objects. All the features of frame $\bm{P}_{t+1}$, $\bm{P}_{t+3}$, $\bm{P}_{t+5}$, $\bm{P}_{t+7}$, $\bm{P}_{t+9}$ are propagated from the I-frame $\bm{I}_t$ by utilizing motion vectors, rather than using heavy computational network. Moreover, the bounding boxes location and recognition results are reasonable by the guidance of motion cues, and sometimes even exceed the flow estimation results.

For flow estimation, the motion information is more smooth. It has superior performance when the object is small and unclear. But this model composes dozens of convolutional layers. For each neighboring frame, it should calculate the FlowNet first, which seems not elegant. 

To sum up, FlowNet is capable of building detailed motion information. And the motion vector can quickly provide the motion cues which help to speed up detection. This comparison shows the potential of compressed video based detection methods. It fully exploits the codec information and makes the model more elegant.

\subsection{Comparison with state-of-the-art systems}
In this section, we show the runtime speed and performance of the related methods in Table \ref{table2} and Figure \ref{fig7}. In Table \ref{table2}, the methods are divided into three groups: single frame baseline \cite{dai2016r}, box-level and feature-level propagation. We also present the detailed accuracy-runtime tradeoff of baseline methods whose performance is above 70\% mAP in Figure \ref{fig7}. And the runtime includes the cost of data preprocessing.

\begin{table*}[!t]
\begin{minipage}{0.5\textwidth} 
\centering 
\renewcommand\arraystretch{1.2}
\begin{tabular}{p{2cm}|p{2.8cm}|p{2.5cm}}
\hline
Methods&Method&mAP(\%)\\
\hline
Single Frame&R-FCN\cite{dai2016r}&73.6\\
\hline
\multirow{4}*{\shortstack{Box\\Propagation}}&ST-Lattice\cite{chen2018optimizing}&77.8\\
&TCNN\cite{kang2017t}&73.8\\
&Seq-NMS \cite{han2016seq} & 52.2\\
&TCN\cite{kang2016object}&47.5\\
\hline
\multirow{6}*{\shortstack{Feature\\Propagation}}&MANet\cite{wang2018fully}&78.1\\
&FGFA\cite{zhu17fgfa}&76.5 \\
&DFF\cite{zhu17dff}&73.1\\
&TPN\cite{kang2017tpn}&68.4\\
&Mobile\cite{liu2017mobile}&54.4 \\
&\textbf{Ours (MMNet)}&73.0(41fps) $\sim$76.4(10fps)\\
&\textbf{Ours (+PostProc.)}&74.8(55fps) $\sim$79.2(8fps)\\
\hline
\end{tabular}  
\caption{Performance comparison with state-of-the-art systems on the ImageNet VID validation set. The mean average precision (in \%) over all classes is shown.}
\label{table2}
\end{minipage}%
\hspace{0.24in}
\begin{minipage}{0.44\textwidth} 
\centering 
\includegraphics[width=0.9\textwidth]{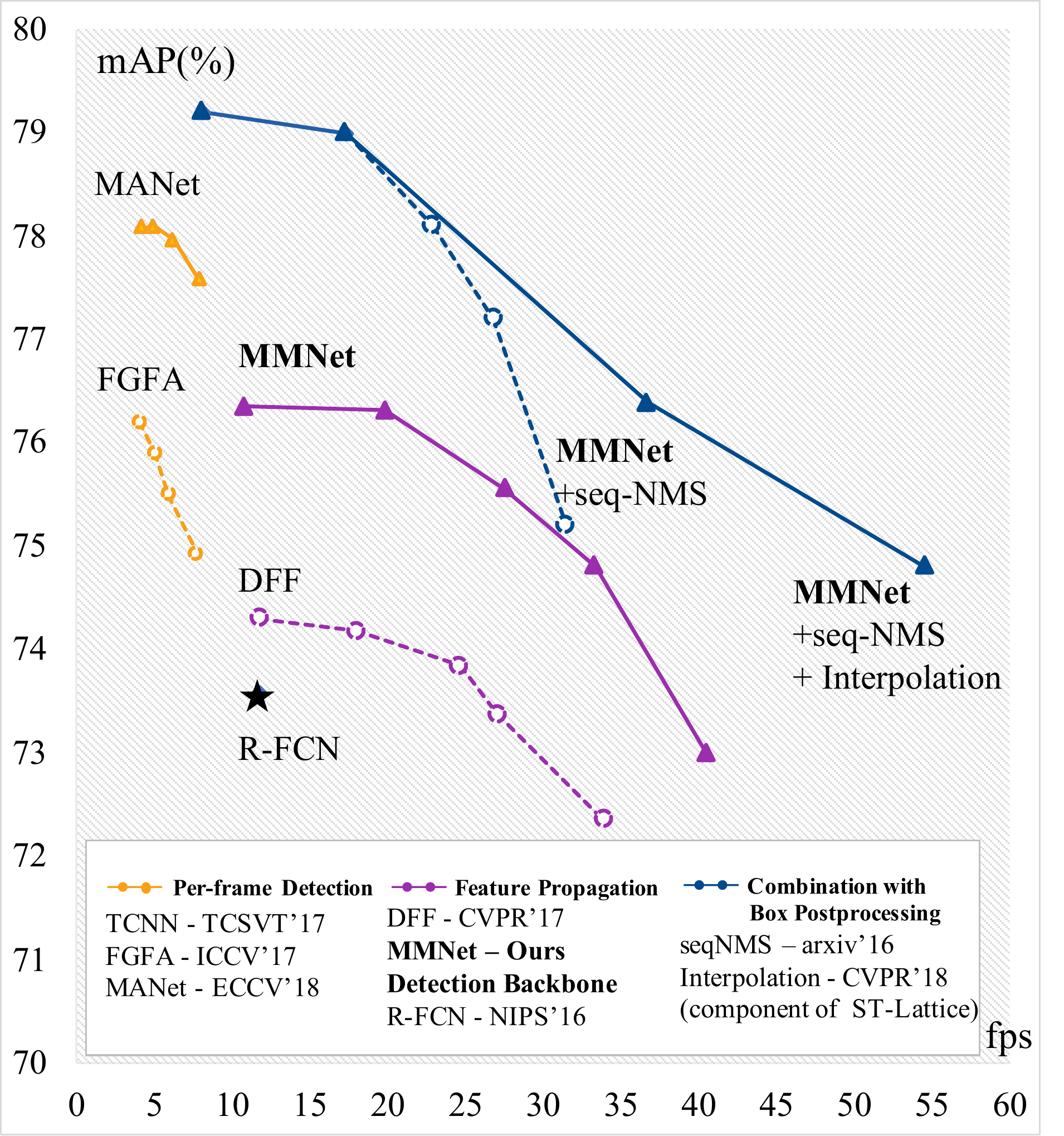} 
\captionof{figure}{The detailed speed and accuracy of some typical methods. The runtime is measured on an NVIDIA Titan X Pascal GPU.}
\label{fig7} 
\end{minipage}%
\end{table*}

From the comparison in Figure \ref{fig7}, we find that:

\noindent \textbf{Per-frame detection (yellow)}: MANet\cite{wang2018fully} has the best performance among these previous works, whereas it takes about 260ms to detect for one frame. All of these per-frame detectors use heavy computational networks ($<$10fps);

\noindent \textbf{Feature propagation(purple)}: After producing features on a keyframe, DFF \cite{zhu17dff} propagates features by using flow estimation. Compared with DFF, our model achieves better performance on both accuracy and runtime speed.

\noindent \textbf{Box postprocessing(blue)}: The box-level propagation is complementary with feature-level propagation. We select two typical methods seq-NMS\cite{han2016seq} and interpolation (part of ST-Lattice) \cite{chen2018optimizing} as baselines. When we combine them with our MMNet, they steadily push forward the performance envelope. 

\noindent \textbf{To sum up}, the MMNet performs well in both accuracy and speed, and it can be easily incorporated with box-level post-processing methods.

\section{Conclusions}
In this paper, we propose a fast object detection model incorporating motion-aided memory network called MMNet. It can be directly applied to compressed videos. Different from the previous work, we use motion information stored and transmitted within a video stream, rather than building another model to retrieve motion cues. We use the I-frame as the reference frame, and explore the memory network to transfer features to next P-frames. All these operations are designed with respect to compressed videos. Compared to ``Action Recognition'' which uses MVs as an extra input stream to enhance global video-level information, the MMNet adopts MVs to predict detailed per-frame features for ``Video Object Detection'' in a fast and accurate manner.  We conduct extensive experiments, including ablation study, visualization and performance comparison, demonstrating the effectiveness of the proposed model.


{\small
\bibliographystyle{ieee_fullname}
\bibliography{egbib}
}

\end{document}